\documentclass[review]{elsarticle}
\usepackage{float}
\usepackage{amsmath}
\usepackage{nicealgo}
\usepackage{multirow}
\usepackage{color}

\usepackage{hyperref}

\journal{ISPRS Journal of Photogrammetry and Remote Sensing}

\usepackage{xcolor}

\begin{document}

\begin{frontmatter}

\title{Correcting rural building annotations in OpenStreetMap using convolutional neural networks}


\author[unicamp]{John E. Vargas-Mu\~{n}oz}
\author[wur]{Sylvain Lobry}
\author[unicamp]{\\Alexandre X. Falc\~{a}o}
\author[wur]{Devis Tuia}

\address[unicamp]{Laboratory of Image Data Science, Institute of Computing, University of Campinas, Campinas, Brazil}
\address[wur]{Laboratory of Geo-information Science and Remote Sensing, Wageningen University \& Research, the Netherlands}

\begin{abstract}
\textbf{This is the pre-acceptance version, to read the final version published in the ISPRS Journal of Photogrammetry and Remote Sensing, please go to: \url{https://doi.org/10.1016/j.isprsjprs.2018.11.010}}

Rural building mapping is paramount to support demographic studies and plan actions in response to crisis that affect those areas. Rural building annotations exist in OpenStreetMap (OSM), but their quality and quantity are not sufficient for training models that can create accurate rural building maps. The problems with these annotations essentially fall into three categories: (i) most commonly, many annotations are geometrically misaligned with the updated imagery; (ii) some annotations do not correspond to buildings in the images (they are misannotations or the buildings have been destroyed); and (iii) some annotations are missing for buildings in the images (the buildings were never annotated or were built between subsequent image acquisitions). First, we propose a method based on Markov Random Field (MRF) to align the buildings with their annotations. The method maximizes the correlation between annotations and a building probability map while enforcing that nearby buildings have similar alignment vectors. Second, the annotations with no evidence in the building probability map are removed. Third, we present a method to detect non-annotated buildings with predefined shapes and add their annotation. The proposed methodology shows considerable improvement in accuracy of the OSM annotations for two regions of Tanzania and Zimbabwe, being more accurate than  state-of-the-art baselines.
\end{abstract}


\end{frontmatter}


\section{Introduction}
\label{sec:introduction}

The amount of publicly available mapping information in web services, like Google Maps and OpenStreetMap (OSM), is large, covering great part of the existing human settlements in the world. Although mapping information of buildings and several other man-made structures are largely available for urban areas, a significant amount of rural buildings is not mapped in any of the aforementioned systems. Rural building mapping information is important to assist demographic studies and help Non-Governmental Organizations to plan actions in response to crises~\footnote{\url{https://www.missingmaps.org/}}. There is therefore a need for creating (or at least updating) urban footprint vector databases in rural areas.

Several works in the literature have approached this problem as the one of detecting buildings in remote sensing images using shape, color, edge, and texture knowledge-based features~\cite{Sirmacek_2009, Benarchid_2013}. More recently, Convolutional Neural Networks (CNNs, for a review in remote sensing see~\cite{Zhu17}) in combination with other image processing methods have been used to detect and delineate buildings in urban areas with successful results~\cite{Mnih_2012, Maggiori_2017cnn, Saito_2016}. Most commonly, the pixel (or region) level detections are merged into vectorial shapes in a post-processing step. In~\cite{Marcos2018cvpr}, a CNN model was proposed to avoid this postprocessing step: vector footprints of buildings are learned directly, by defining the building outline definition as an active contour model, whose parameters are learned with a CNN. The investigation of building detection using deep learning is a field of growing interest, also supported by recent data processing competitions in this direction, e.g. DeepGlobe~\cite{Dem18}.

Irrespectively of the strategy chosen, the main drawback of using CNN methods in remote sensing is the need of large amount of labeled data for training. In recent research, OSM annotations have been used as repositories of large labeled data collections. In GIScience, this source of data has proven to be very powerful, and several works have proposed methods to automatically predict attributes of OSM objects. For example in~\cite{Jilani_2016prob}, the authors proposed a methodology for automatic prediction of street labels (e.g., motorway and residential). In~\cite{Fan_2014}, authors proposed a method using geometrical properties of the OSM annotation polygons to predict the types of buildings (e.g., residential, industrial and commercial). In~\cite{Fleischmann_2017navigation}, OSM data was used to improve robot navigation for autonomous driving and in~\cite{Wang_2017} OSM data was used for 3D building modeling, allowing visualization of indoor and outdoor environments in 3D maps. Authors in~\cite{Srivastava_2018} use Google Street View pictures to predict the landuse of the footprints. They use OSM annotations as labels to train a deep learning model. Within the remote sensing building segmentation field, OSM annotations of urban areas have been recently used in~\cite{Audebert_2017} and~\cite{Kaiser_2017} as label information to perform semantic segmentation of buildings and roads. The INRIA building detection challenge uses corrected OSM footprints as labels~\cite{maggiori2017dataset}. 

Despite the appeal of using OSM data for training deep learning models, the quality of these data is uneven.  
Usually CNNs trained with this type of reference data can learn to predict the location of the object but not the exact object extent~\cite{Mnih_2012}. 
Several works proposed methods that can be useful to improve the quality of the OSM data, both for attribute classification and positional inaccuracies. Authors in~\cite{Basiri_2016} detect errors in OSM annotations of roads using patterns extracted from GPS tracking data. For instance, indoor corridors wrongly classified as tunnels can be detected using tracked trajectories of cars and pedestrians. 
In~\cite{Hashemi_2015}, distance, directional, and topological relationship of OSM objects are used to detect inconsistencies. 

\begin{figure}[!t]
\begin{center}
\begin{tabular}{cc} 
\multicolumn{2}{c}{}\\
  \includegraphics[width=0.38\columnwidth]{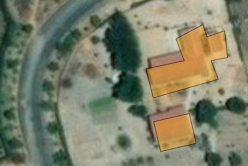} & 
  \includegraphics[width=0.4\columnwidth]{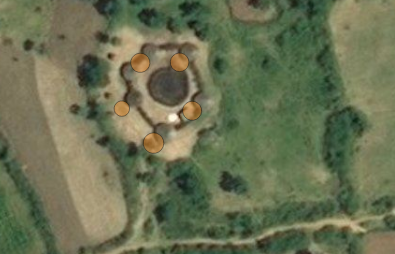} \\
  (a) Urban buildings & (b) Rural buildings
\end{tabular} 
\end{center}
\caption{Misaligned OSM building annotations (in orange) superimposed on the imagery obtained from Bing maps: a) For urban building misaligned annotations, there is a considerable overlap with the object in the imagery; b) For the case of rural building misaligned annotations, some buildings in the imagery and their corresponding annotations do not overlap.\label{fig:rural_building_example1}}
\end{figure}

OSM has gathered and made publicly available large amounts of building annotation data. But if the quality of OSM data has been judged sufficient for urban areas~\cite{Est13}, the same does not hold in rural areas, especially because of the lower update rate and the drop in the number of volunteers out of cities. By analyzing  available OSM data in rural areas, we observed that the annotations performed by the volunteers suffer from three main issues, mostly due to infrequent imagery updates and incomplete/inaccurate volunteer annotations~\cite{Basiri_2016, Barron_2014}:

\begin{enumerate}
\item[(a)] the locations of building annotations might be inaccurate: building footprints are often present, but displaced on the image plane by up to $9$ meters. These displacements are often due to the fact that the image used to digitize the footprint does not correspond to the image being used for analysis. Two examples of misalignments are given in Figure~\ref{fig:rural_building_example1}; 
\item[(b)] some annotations do not correspond to buildings in the imagery: in this case, some buildings might have been demolished, or simply the annotations by the volunteers are erroneous~\cite{Fonte15}; 
\item[(c)] some objects that appear in the imagery are not present in the annotation dataset: in this case, some buildings might have been missed by the volunteers or new buildings might have been built in between the two image acquisitions. 
\end{enumerate}
 
In order to deal with inaccurate reference building
data, the authors in~\cite{Mnih_2012} propose a loss function to reduce the effect of this problem, while the authors in~\cite{Maggiori_2017rnn} use a Recurrent Neural Network to improve the classification maps with a small set of perfectly and manually annotated data. However, as mentioned above, for rural buildings the problem of inaccurate annotations is more severe, since buildings are smaller and scarcer than urban buildings in OSM~\cite{Chen_2017}. As one can see in Figure~\ref{fig:rural_building_example1}, there exists considerable overlapping areas between urban buildings and the misaligned OSM annotations, while some rural buildings in the image and the OSM annotations do not overlap.

In this work, we propose a methodology to correct OSM rural building annotations. We tackle the three problems above simultaneously, with a three-stage strategy based on the predictions of a fully convolutional deep learning model that estimates the likelihood of presence of buildings.

\begin{enumerate}
\item[(i)]First, we propose a method to align buildings and their annotations based on Markov Random Field (MRF)~\cite{Besag_86}. We make the hypothesis that alignment errors can be fixed by simply translating the annotations themselves, since we observed that this type of error is the most frequent in rural areas (see Figure \ref{fig:rural_building_example1}(b)). MRFs have been successfully applied to solve registration problems in several image domains~\cite{Glocker_2011, Marcos_2016,Vak16}. Our MRF-based method maximizes the correlation between OSM annotations and a predicted building probability map, while enforcing that nearby buildings have similar alignment vectors (shift correction vectors). Usually, rural buildings appear in small groups with the same alignment errors (as a given area is annotated on one image by the same volunteer, this whole area will present a similar misalignment when the imagery is updated). For this reason we use nearby rural buildings as nodes of a small MRF graph. The method then computes a single  alignment vector for all the buildings in each small group of rural buildings. 

\item[(ii)]Second, the OSM annotations with no evidence in the previously computed building probability map are removed. For each OSM annotation, we compute the mean building probability value of the pixels contained in the aligned annotations. If the computed values are smaller than a threshold~\cite{Glasbey_1993}, we remove the OSM annotations.

\item[(iii)]Third, we present a CNN-based method for adding new building annotations. Since the variety of rural building shapes and sizes is very small as compared to the ones of urban buildings, the CNN estimates one of 18 commonly appearing rural building shapes for each non-annotated building. 
\end{enumerate}

In Section~\ref{sec:methodology} we present the proposed methodology to correct OSM rural building annotations. Section~\ref{sec:data_and_experimental_setup} shows the dataset and the setup of our experiments and Section~\ref{sec:results} compares the results of our proposed method with other baseline methods. Section~\ref{sec:conclusions} concludes the paper.

\begin{figure}[H]
\centering
\includegraphics[width=0.95\columnwidth]{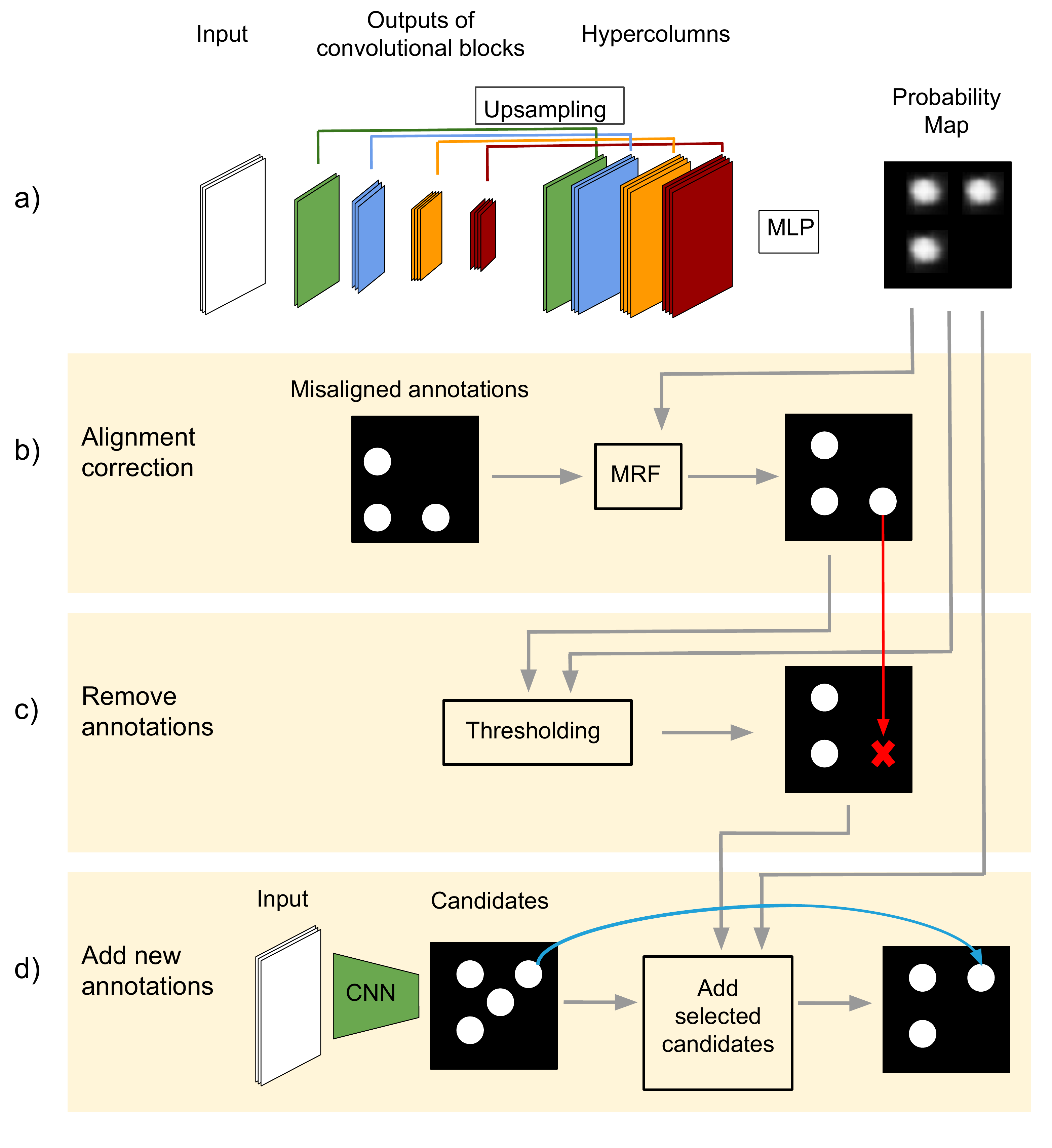}
\caption{Proposed methodology to correct OSM rural building annotations: a) predict a building probability map from an aerial image using a CNN trained for per-pixel classification; b) correct alignment errors in the OSM annotations using a MRF-based method and a building probability map; c) remove OSM annotations based on the aligned annotations, a building probability map, and a thresholding method; d) add new annotations selected from a set of candidates obtained by a CNN that predicts rural buildings with predefined shapes.\label{fig:methodology}}
\end{figure}

\section{Methodology}
\label{sec:methodology}

Our methodology to correct OSM annotations of rural buildings requires a fully convolutional neural network (CNN) model trained to generate a building probability map for the overhead image (Figure~\ref{fig:methodology}a): this method is detailed in Section~\ref{subsec:compute_probmap}. Once this classifier is trained, the building correction module consists of three main tasks, as described in Section~\ref{sec:introduction}. Figures~\ref{fig:methodology}b-d illustrate them, from top to bottom. In sections~\ref{subsec:align_annotations} to~\ref{subsec:add_annotations} we detail these methods. 

\subsection{Computing building probability maps}
\label{subsec:compute_probmap}

In order to correct OSM rural building annotations, we use a building probability map obtained by a CNN model that performs pixel classification. In this work we use a CNN model based on~\cite{Volpi_2017} that is trained on a small set of manually verified/corrected rural building OSM annotations. The CNN model performs four convolutional blocks (convolution followed by spatial pooling, non-linear activation and batch normalization operations) but, differently from~\cite{Volpi_2017} that uses deconvolutions to upsample the feature map, we  apply the concept of hypercolumns~\cite{Hariharan_2015} to perform pixel classification. We modified the original hypercolumn model in the same way as for the baselines of~\cite{Marcos_isprs_2018}: the hypercolumns are obtained by upsampling the outputs of previous convolutions to the size of the input image using bilinear interpolation. This makes the training of the CNN more efficient and with similar performance. These activations are then stacked to a single tensor which is used to train a Multi-layer Perceptron classifier to perform pixel classification. The architecture of the described CNN is presented in Figure~\ref{fig:methodology}a, while the details of the specific architecture are presented in Section~\ref{sec:data_and_experimental_setup}. 

\begin{figure}[!t]
\begin{center}
	\includegraphics[width=0.5\columnwidth]{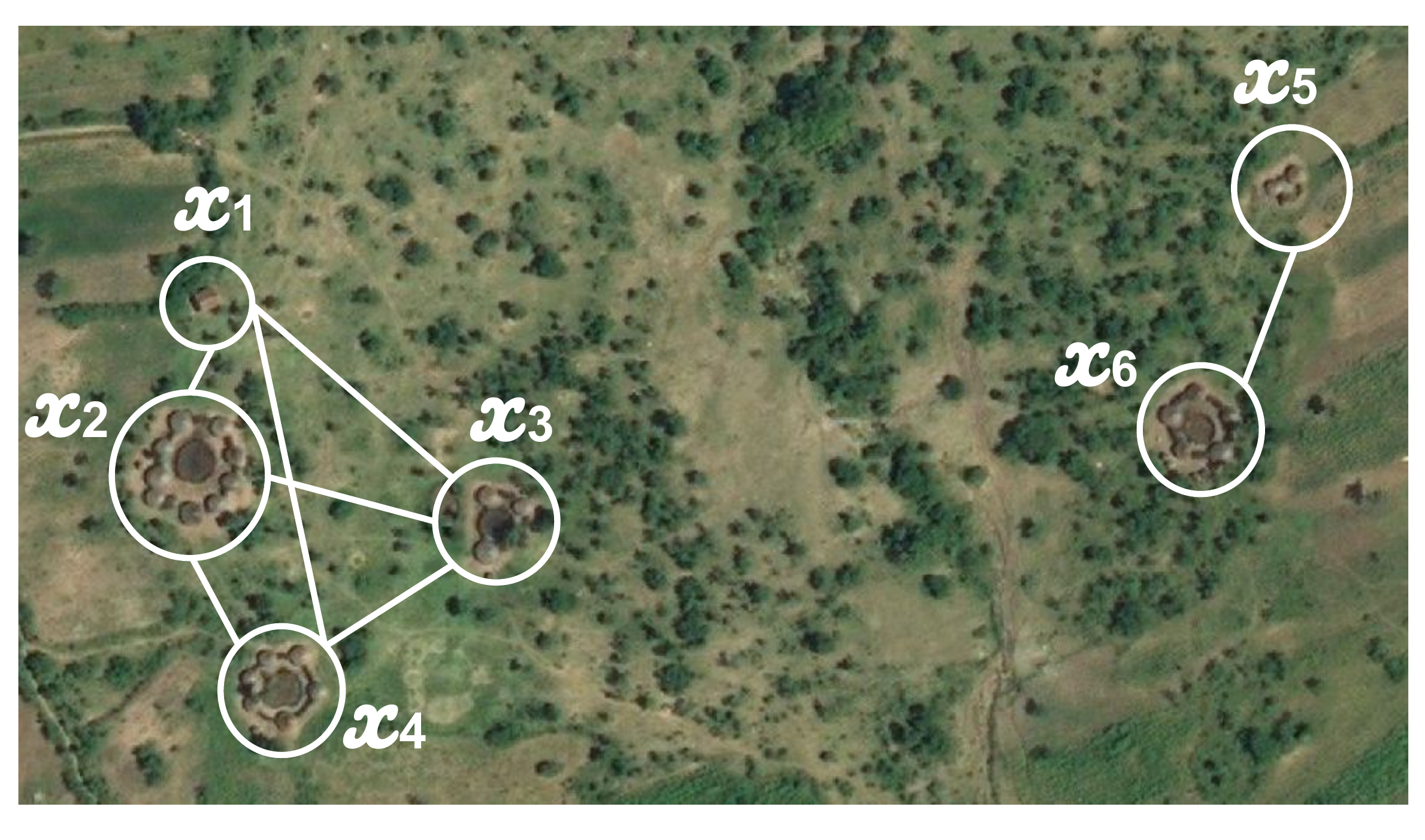}
	\caption{Neighboring system of the proposed MRF method. Groups of rural buildings are used as nodes of the MRF graph.
	\label{fig:mrf_graph}} 
\end{center}
\end{figure}

\subsection{Aligning OSM rural building annotations}
\label{subsec:align_annotations}
The building registration problem is considered as the problem of aligning the vector shapes from OSM to the predictions of the CNN (Figure~\ref{fig:methodology}b). Such alignment is performed by estimating alignment vectors, basically shifting every OSM polygon to an area of high building probability in the CNN map.

In order to compute these alignment vectors, we need to measure how well a given shift performs. To this end, we use the correlation between the aligned annotations and the building probability map obtained previously using the image on which the annotations need to be registered. 
Making the hypothesis that rural buildings are gathered in small groups where each building has the same misalignment error, we align groups of buildings instead of individual buildings. This reduces greatly the computational load and is numerically more effective (see the results Section~\ref{sec:results}). Moreover, using groups of buildings instead of single ones makes the results less dependent on the quality of the building probability map.

Additionally, we observed that nearby groups of buildings have similar registration errors. Based on this observation, we build our building registration module on a MRF model using this prior together with the evidence provided by the building probability map. Our method aims at finding the alignment vectors $\mathbf{d} = \{d_0, d_1, \ldots, d_n\}$ that need to be applied to the annotation locations $\mathbf{x}$ based on the a probability map $\mathbf{y}$. Groups of buildings, or \emph{sites}, are used as nodes of the MRF graph (See Figure~\ref{fig:mrf_graph}), where sites $i$ and $j$ are neighbors (i.e., $j \in N_{i}$) in the graph if they are spatially close (see Section~\ref{subsec:model_setup} for more details on the MRF graph definition).

In our MRF formulation, the unary term is obtained by using the normalized correlation $C(d_i(x_i), y_i)$ between the annotation after alignment $d_i(x_i)$ and the building probability map $y_i$. This term is equal to the average of the predicted probability values $y_i$ of the pixels contained in the aligned annotation $d_i(x_i)$. The pairwise term is defined by the dissimilarity (vector norm of the difference of two vectors) between the alignment vector $d_i$ of the annotation $i$ and the alignment vectors $d_j$ of neighboring annotations $j \in N_i$~\cite{Marcos_2016}.  
The optimal set of alignment vectors $\mathbf{\hat{d}}$ for the annotations is defined by:

\begin{align}
\mathbf{\hat{d}} &= \arg\min\limits_\mathbf{d\in\mathcal{D}^\mathcal{N}} \sum\limits_i U(d_i | x_i, y_i)\label{eq:energy_total}\\
&=\arg\min\limits_\mathbf{d\in\mathcal{D}^\mathcal{N}} \sum\limits_i -\log C(d_i(x_i), y_i) + \beta \sum_{j \in N_i} \frac{1}{Z} ||d_i - d_j||_2,\nonumber
\end{align}

where $\mathcal D = \{ D_1, D_2, \ldots , D_m \}$ is the set of all possible $m$  alignment vectors, $\beta$ is the spatial regularization parameter and $Z$ is a normalization factor, defined as the maximum possible distance between two alignment vectors in $\mathcal D$. To compute the optimal $\mathbf{\hat{d}}$ by minimizing the energy function $U$, we use the Iterative Conditional Modes (ICM)~\cite{Besag_86} algorithm initialized with $d_i~=~\arg\max\limits_{d\in\mathcal{D}}C(d(x_i), y_i)$. As this initialization is already a good heuristic (see Section \ref{sec:results}), the ICM algorithm allows to obtain a good solution in a few iterations. The inclusion of a distance-based weight in the pairwise term does not lead to better performances, so it is omitted for clarity. We presented preliminary results of our proposed method for alignment of OSM annotations in the conference paper~\cite{VargasMunoz_2018}. Algorithm~\ref{alg.alignment_algo} summarizes the proposed method for aligning OSM annotations.

\begin{figure}[t]
\centering
\begin{minipage}{.8\linewidth}

\begin{nicealgo}{alg.alignment_algo}
\naTITLE{MRF-based alignment algorithm} \naPREAMBLE \naINPUT{Original OSM annotations $M$ and building probability map $\mathbf{y}$}  
\naOUTPUT{Alignment vectors $\mathbf{d}$}  
\naBODY 

\na{Group the original rural building annotations $M$ according to their\\ ${}$\hspace{1.8em} spatial distance from each other, obtaining the set of building groups {$\mathbf{x}$}.}

\na{Define the neighbors $N_i$ of each site $i$ as spatially close sites.}

\na{Initialize each $d_i$ to $\arg\max\limits_{d\in\mathcal{D}}C(d_i(x_i), y_i)$}

\na{Run Iterated Conditional Modes (ICM) for $MaxIters$ iterations}
  
\naBEGIN{\naFOR $t \leftarrow 1 ... MaxIters$, \naDO}

\naBEGIN{\naFOREACH $x_i \in \mathbf{x}$, \naDO}

\naBEGIN{\naFOREACH $D \in \mathcal D$, \naDO}

\na{Compute energy $U(D | x_i, y_i)$, equation~\eqref{eq:energy_total}}

\naBEGIN{\naIF $U(D | x_i, y_i) < U(d_i | x_i, y_i)$, \naTHEN}

\naENDN{4}{$d_i \leftarrow D$}

\na{Return $\mathbf{d}$}

\end{nicealgo}

\end{minipage}
\end{figure}

\subsection{Removing incorrect building annotations}
\label{subsec:remove_annotations}

In order to remove OSM annotations that no longer exist in the updated imagery (Figure~\ref{fig:methodology}c), we compute the mean building probability value of the pixels contained in the aligned annotations. We observe that the histogram of these average probability values roughly follows a bimodal distribution. The group of annotations close to the first local maximum corresponds to some of the few annotations that have average probability values close to zero (showing high evidence that there is no longer a building in that location of the imagery) while the other group of annotations gathered around the second and most prominent local maximum corresponds to the majority of the aligned annotations that have higher average probability values. Since Otsu's thresholding method~\cite{Otsu_1979} is known not to perform well for unbalanced distributions~\cite{Lee_1990} we use the Minimum threshold method~\cite{Glasbey_1993}. This method iteratively smooths the histogram until only two local maxima are found. After that, the minimum value between the two local maxima is selected as the threshold. We then remove annotations, which have an average probability value below this threshold.

\subsection{Add new building annotations}
\label{subsec:add_annotations}

\begin{figure}[!t]
\begin{center}
	\includegraphics[width=0.95\columnwidth]{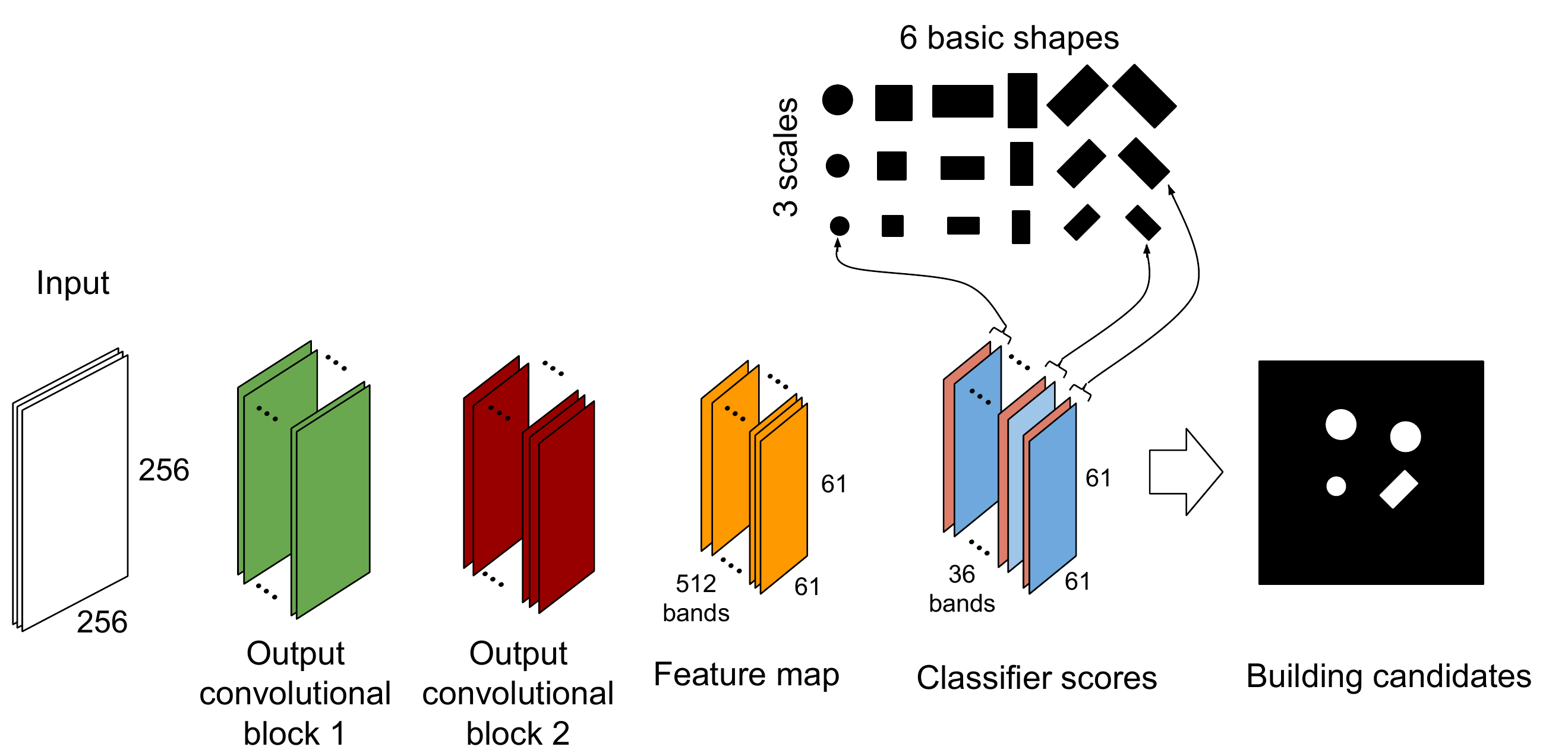}
	\caption{CNN model for adding new annotations of buildings that appear for the first time in the updated imagery.
	\label{fig:cnn_detection}} 
\end{center}
\end{figure}

The last task is the addition of new building footprints (Figure~\ref{fig:methodology}d). We observed that rural buildings appear with very few different shapes in the imagery (e.g., circles and rectangles), as compared to urban buildings. Therefore, we make the hypothesis that a restricted number of shapes is sufficient to represent most buildings in rural areas.
Inspired by this, we compile a set of 18 commonly appearing shapes and propose a CNN model that predicts if a building with one of these predefined shapes is present in a particular location of the imagery (see Figure~\ref{fig:cnn_detection}). Based on our observations, we select 6 basic geometrical shapes: a circle of radius 3.3 meters, a square of side 4.8 meters, a rectangle of sides 3.6 and 6 meters, and the same rectangle rotated by $45^\circ, 90^\circ$ and $135^\circ$. Furthermore, for each base shape we generate two more scaled versions, by approximately increasing its area by a factor of 2 and 4, resulting a total of 18 considered shapes (see Figure \ref{fig:cnn_detection}). 

The architecture of the proposed CNN model is depicted in Figure~\ref{fig:cnn_detection}: we apply two convolutional blocks followed by one convolutional layer to the input image of size $256 \times 256$, leading to a $61 \times 61$ feature map with 512 activations per location (details of the specific architecture are presented in Section~\ref{sec:data_and_experimental_setup}). 
Afterwards, we apply a $1 \times 1$ convolutional layer that outputs a matrix of size $61 \times 61$ and $36$ bands. This operation is performed to compute scores for the two classes of interest (presence or absence of buildings) with the 18 different shapes in each location of the $61 \times 61$ grid. This means that we have a different classifier for every building shape. Every pixel in the $61 \times 61$ grid corresponds to one location in the original $256 \times 256$ input image. Therefore, the location of our building predictions will have an additional approximation error of less than four pixels. 

For training the CNN model, we use a cross entropy loss on the sum of the binary shape classification problems. We consider as positive samples of a given building shape, rural buildings with more than $0.75$ Intersection over Union (IoU) value with a shape mask. The rural buildings with less than $0.30$ IoU value with a shape mask are considered as negative samples for that particular building shape. The threshold values are chosen empirically based on the object detection method presented in~\cite{Ren_2017}. Note that if we choose a higher value for the positive sample's threshold, we might ignore some buildings that have very similar desired shape and if we use lower values for that threshold, we would take the risk of including buildings whose shape does not fit with the desired building shape.

The output of this CNN model is a set of rural building candidates that have predefined shapes. We select a subset of these candidates based on the building probability map and the aligned building annotations, obtained after the annotation removal process. We filter out all the candidates that have average probability values (as obtained by the CNN model that performs per-pixel classification) and detection probability values (obtained by the CNN model described in this section) lower than a certain threshold $t$. In case of overlapping candidates, we select the one with the highest sum of average probability and detection probability values.

\section{Data and experimental setup}
\label{sec:data_and_experimental_setup}

\subsection{Datasets} We evaluate our method with OSM rural building data from two different countries, namely the United Republic of Tanzania and the Republic of Zimbabwe. The evaluation data collected from these two countries have different characteristics: while the Tanzania's evaluation region contains severe misaligned and incomplete annotations, the evaluation region in Zimbabwe contains more accurate annotations. The Bing imagery utilized for the two datasets were acquired between $2004$ and $2014$, while the annotations obtained from OSM were performed by volunteers between $2013$ and $2018$. Bing maps provides an API to obtain aerial imagery (red, green and blue channels) at different spatial resolutions (e.g., 119 cm, 60 cm, 30 cm). In this work, for the training and testing datasets, we use Bing maps imagery of 30 cm spatial resolution since we wanted to obtain accurate building classification maps with the CNN. The lower the spatial resolution, the higher are the chances to obtain inaccurate building classification maps, with missing buildings and false positives. 
Therefore, we recommend the use of imagery with 60 cm or higher spatial resolution  that can be obtained from pansharpened images of  satellites such as QuickBird, GeoEye, Pl\'{e}iades, WorldView-2,  WorldView-3, and WorldView-4.

In order to train the CNN model that predicts the building probability maps (Section~\ref{subsec:compute_probmap}), we use $3134$ OSM rural buildings annotations. These OSM annotations were manually verified/corrected on a set of Bing aerial images, that cover $23.75$ km$^2$, 
acquired over the Geita, Singida, Mara, Mtwara, and Manyara regions of Tanzania. In order to obtain the building probability maps for the Zimbabwe dataset, we finetune the CNN model trained on Tanzania's annotations with a small dataset of $559$ building annotations obtained from the region of Matabeleland North in Zimbabwe.

In order to evaluate our methodology, we create validation datasets spatially disconnected from the training regions. The first one is composed of $1094$ manually corrected misaligned building annotations located close to the city of Mugumu in Tanzania, where we found OSM annotations with different misalignment orientations. The second dataset is composed of $811$ manually corrected misaligned annotations located in the region of Midlands in Zimbabwe. The validation dataset from Tanzania consists of three rural areas, for which we obtained Bing images of sizes (in pixels) $7936 \times 8192$, $8192 \times 8192$ and $7168 \times 3840$, respectively. The validation dataset from Zimbabwe consists of four rural areas that were covered by Bing images of sizes $4096 \times 3328$, $4096 \times 3584$, $5120 \times 4352$ and $5120 \times 4352$ pixels, respectively.\\

\subsection{Model setup and evaluation procedures}
\label{subsec:model_setup}
\begin{itemize}
\item[-] \textbf{Building probability CNN.} For the CNN model that obtains the building probability maps, the numbers of filters in the convolutional layers are 32, 64, 128, and 128, with corresponding kernel sizes of $7 \times 7$, $5 \times 5$, $5 \times 5$ and $3 \times 3$. We apply max-pooling (with stride $2$ and kernel size $3 \times 3$), Rectified Linear Unit
(ReLU) as activation function and batch normalization after every convolutional layer. We use $90\%$ rate dropout after on the final fully connected layer. We train the model for $5000$ stochastic gradient descent iterations using a learning rate of $0.001$ and other $5000$ iterations using a learning rate of $0.0001$.
\item[-] \textbf{MRF graph.} As mentioned in Section~\ref{subsec:align_annotations}, we use groups of buildings as nodes of the MRF graph. A building belongs to a group if its center is less than 21 meters away from the center of any of the buildings in this group. 
In the MRF graph, every group of buildings is then connected to the 5 closest groups of buildings. 
Both parameters (minimum distance to single buildings for inclusion and number of closest groups) have been set empirically.

\item[-] \textbf{Alignment with MRF.} The alignment vectors $\mathcal{D} = \{(x,y), x \in \mathcal{D}_x, y \in \mathcal{D}_y\}$ are defined with $\mathcal{D}_x = \mathcal{D}_y = \{-30, -29, ...0, ..., 29, 30\}$ (values in pixels) based on the maximum expected misalignment. We set the MRF spatial regularization parameter $\beta = 2$ and the maximum number of iterations of the ICM to $10$, experimentally. The ICM algorithm has converged before the tenth iteration in all the datasets.
\item[-] \textbf{Building generation by CNN.} For the CNN model that detects buildings with predefined shapes, the number of filters in the convolutional layers were, 32, 128, 512 and 16 with corresponding kernel sizes of $5 \times 5$, $3 \times 3$, $3 \times 3$ and $1 \times 1$. We apply max-pooling (with stride $2$ and kernel size $3 \times 3$), ReLU as activation function and batch normalization after the first two convolutional layers. In order to select the predicted building candidates to be added to the OSM annotations, we use a threshold value $t = 0.80$, that was found experimentally. This high threshold value is selected to decrease the number of false positives. 
\end{itemize}

We evaluated the performance of the proposed method using the Precision, Recall and F-score metrics with a pixel-level evaluation of the predictions.

\section{Results}
\label{sec:results}

We compare the proposed method  for alignment of OSM annotations (\texttt{MRFGroups}) with the original annotations (`\texttt{without alignment}') and the following competitors from the literature:
\begin{itemize}
\item[-] \texttt{`DeformableReg'}, a deformable registration method trained using an unsupervised approach that optimizes a similarity metric between pairs of images~\cite{deVos_2017deformable_registration}. \texttt{DeformableReg} analyzes pairs of image patches extracted from building classification maps and OSM annotation maps (of the training set) to generate a displacement vector field. During the inference phase, these vectors allow to perform the registration of the OSM annotation maps into the building classification maps for the test dataset. \texttt{DeformableReg} is trained for $10000$ stochastic gradient descent iterations using a learning rate of $0.0001$. 
\item[-] \texttt{`Semantic segmentation'}, the fully convolutional CNN-based segmentation model in~\cite{Maggiori_2017cnn}. The CNN architecture is composed of several convolutional layers that extract features followed by a deconvolutional layer that output the final per-pixel classification. The model is trained for $5000$ stochastic gradient descent iterations using a learning rate of $0.001$ followed by $5000$ iterations using a learning rate of $0.0001$.
\end{itemize}
In addition to the competitors from the literature, we report results obtained by our model in varying conditions: 

\begin{itemize}
\item[-] \texttt{`CorrBuildings'}. When selecting the alignment vectors that maximize the correlation between individual building annotations and a building probability map (equivalent to use our MRF alignment model with $\beta = 0$). 
\item[-] \texttt{`CorrGroups'}. When obtaining the alignment vectors that maximize the correlation between groups of buildings and the building probability map. 
\item[-] \texttt{`MRFBuildings'}. When performing the alignment with the proposed MRF formulation, but using individual buildings as nodes of the MRF graph. It is mostly meant to assess computational speedups and the loss of precision when using individual of buildings.
\item[-] \texttt{`AbsDifference'}. When obtaining the alignment vectors that minimize the sum of absolute difference between groups of buildings and the building probability map. 
\item[-] \texttt{`MutualInfo'}. When obtaining the alignment vectors that maximize the mutual information between groups of buildings and the building probability map.

\end{itemize}

\begin{table}[!t]
\centering
\caption{Pixel-based performance of alignment correction methods for the Tanzania evaluation dataset.}
\label{tab:results_tanzania}
\vspace{1em}
\begin{tabular}{|p{5.3cm}|r|r|r|r|}
\hline
\multirow{2}{*}{Methods} & \multirow{2}{*}{Precision} & \multirow{2}{*}{Recall} & \multirow{2}{*}{F-score} & Time \\
& & & & (sec) \\
\hline\hline
{\texttt{Without alignment}} & 0.108 & 0.115 & 0.111 & 0  \\ 
\hline
{\texttt{CorrBuildings}} & 0.565 & 0.460 & 0.507 & 141.7  \\ 
{\texttt{CorrGroups}}& 0.620 & 0.658 & 0.639 & 102.3 \\ 
{\texttt{MRFBuildings}} & 0.644 & 0.644 & 0.644 & 147.8 \\ 
{\texttt{AbsDifference} }& 0.303 & 0.322 & 0.312 & 41.2 \\
{\texttt{MutualInfo} }& 0.570 & 0.606 & 0.587 & 520.6\\
\texttt{MRFGroups} (proposed method) & \textbf{0.638} & \textbf{0.677} & \textbf{0.657}  & 103.0 \\
{\texttt{DeformableReg} }~\cite{deVos_2017deformable_registration} & 0.380 & 0.500 & 0.430 & 29.3 \\ 
\hline
\end{tabular}
\end{table}

\begin{table}[!t]
\centering
\caption{Pixel-based performance of alignment correction methods for the Zimbabwe evaluation dataset.}
\label{tab:results_zimbabwe}
\vspace{1em}
\begin{tabular}{|p{5.3cm}|r|r|r|r|}
\hline
\multirow{2}{*}{Methods} & \multirow{2}{*}{Precision} & \multirow{2}{*}{Recall} & \multirow{2}{*}{F-score} & Time \\
& & & & (sec) \\
\hline\hline
\texttt{Without alignment} & 0.526 & 0.519 & 0.523 & 0  \\ 
\hline
\texttt{CorrBuildings} & 0.793 & 0.663 & 0.723 & 84.9  \\ 
\texttt{CorrGroups}& 0.821 & 0.810 & 0.816 & 62.0 \\ 
\texttt{MRFBuildings} & 0.832 & 0.800 & 0.816 & 90.1 \\ 
{\texttt{AbsDifference} }& 0.806 & 0.795 & 0.800 & 36.7\\
{\texttt{MutualInfo} } & 0.815 & 0.804 & 0.809 & 428.2\\
\texttt{MRFGroups} (proposed method) & \textbf{0.830} & \textbf{0.819} & \textbf{0.825} & 63.9 \\
{\texttt{DeformableReg} }~\cite{deVos_2017deformable_registration} & 0.700 & 0.735 & 0.717 & 20.8 \\ 
\hline
\end{tabular}
\end{table}

\begin{table}[!t]
\centering
\caption{Pixel-based and object-based performance of the removal and building addition methods for the Tanzania evaluation dataset.}
\label{tab:results_tanzania_removal_and_addition}
\vspace{1em}
\begin{tabular}{|p{4.5cm}|r|r|r|r|r|}
\hline
\multirow{2}{*}{Methods} & \multirow{2}{*}{Precision} & \multirow{2}{*}{Recall} & F-score & F-score & Time \\
& & & (pixel) & (object) & (sec) \\
\hline\hline
{\texttt{Semantic segmentation}}~\cite{Maggiori_2017cnn} & 0.548 & 0.819 & 0.657 & 0.518 & 80.0 \\
\hline
\texttt{MRFGroups}  &  &  &   &  & \\
 \textbf{+ remove}  & 0.763 & 0.673 & 0.715 & 0.743 & 103.3 \\
 \textbf{+ remove, then add (by shape priors)}   & 0.727 & 0.724 & 0.725 & 0.690 & 284.5\\
\textbf{+ remove, then add (by semantic segmentation)} &  0.649 & 0.776 & 0.707 & 0.719 & 183.8\\ 
\hline
\end{tabular}
\end{table}

\begin{table}[!t]
\centering
\caption{Pixel-based and object-based performance of the removal and building addition methods for the Zimbabwe evaluation dataset.}
\label{tab:results_zimbabwe_removal_and_addition}
\vspace{1em}
\begin{tabular}{|p{4.4cm}|r|r|r|r|r|}
\hline
\multirow{2}{*}{Methods} & \multirow{2}{*}{Precision} & \multirow{2}{*}{Recall} & F-score & F-score & Time \\
& & & (pixel) & (object) & (sec) \\
\hline\hline
{\texttt{Semantic segmentation}}~\cite{Maggiori_2017cnn} & 0.653 & 0.782 & 0.712 & 0.519 & 41.0 \\ 
\hline
\texttt{MRFGroups}  &  &  &   &  & \\
 \textbf{+ remove}  & 0.837 & 0.814 & 0.825 & 0.846 & 64.2  \\
 \textbf{+ remove, then add (by shape priors)}  & 0.833 & 0.817 & 0.825 & 0.841 & 180.1 
\\ 
\textbf{+ remove, then add (by semantic segmentation)} & 0.843 & 0.816 & 0.829 & 0.802 & 105.6
\\ 
\hline
\end{tabular}
\end{table}

\subsection{Numerical results}

Tables~\ref{tab:results_tanzania} and ~\ref{tab:results_zimbabwe} present the performances and processing times of several alignment methods for the Tanzania and Zimbabwe evaluation datasets respectively. 
For the Tanzania dataset (Table~\ref{tab:results_tanzania}), we can observe that the original misaligned annotations poorly match the actual building footprints visible in the image. All the alignment methods drastically improve the performance of the misaligned annotations. MRF-based methods show better performances than methods based only on correlation. This shows that adding the prior knowledge of smoothness of the alignment vectors helps to improve the results. We can also observe that the alignment methods based on groups of buildings are more effective and efficient than the ones based on individual buildings. For the case of the Zimbabwe dataset ( Table~\ref{tab:results_zimbabwe}), the performances of the original misaligned annotations are considerably better than the ones of the Tanzania dataset. As in the Tanzania dataset, all the alignment methods considerably improve the performances of the misaligned annotations and the proposed method based on MRF spatial logic applied on groups of buildings outperforms the other baseline alignment methods, as well as the state-of-art semantic segmentation approach in terms of precision and recall.

Tables~\ref{tab:results_tanzania_removal_and_addition} and~\ref{tab:results_zimbabwe_removal_and_addition} show the performance of the proposed methods for the removal of incorrect annotations and the addition of new annotations in the two datasets. As a starting point, they use the proposed~\texttt{MRFGroups}. In order to evaluate the performance of the methods at the object level we consider that a building is detected if its IoU (Intersection over Union) with the ground truth is greater than 0.5. This value corresponds to a misalignment of 2 pixels (60~cm) in both axes when considering the smallest shape (circle) in our dataset.

In the Tanzania dataset, the removal of incorrect annotations considerably improves the precision of the method while maintaining the recall. When the method that used shape priors for adding new buildings annotations is applied, the recall considerably increases. This is at the cost of a slight decrease in precision because of some false positive predictions. However, the gain in recall is larger in the pixel-level evaluation, which is reflected in the improvement of the F-score. Overall in the Zimbabwe dataset the results of the aligned polygons and the result of removing and adding new polygons to the aligned polygons are equivalent. This happens because most of the buildings in the imagery are already well detected and considerably well delineated by the aligned annotations. Thus, few candidates are removed and new building candidates, as predicted by the proposed CNN, are already at their pre-annotated locations. Therefore, very few new candidate buildings are added.

\subsection{Analysis of shape priors}

In Tables~\ref{tab:results_tanzania_removal_and_addition} and ~\ref{tab:results_zimbabwe_removal_and_addition} we also compare our proposed methods with the fully convolutional semantic segmentation approach proposed in~\cite{Maggiori_2017cnn} (line \texttt{`Semantic segmentation'}).
As it can be observed, in both datasets the proposed methods achieve better performances than this baseline. Alternatively, one could also use a semantic segmentation method (e.g. \cite{Maggiori_2017cnn}) to add new building footprints after running \texttt{MRFGroups} and removing incorrect footprints: this result is reported in the last line of both tables (see \texttt{`+ remove, then add (by semantic segmentation)'}). In this case, we observe  similar numerical performances to our proposed method in terms of F-score. Our proposed method is more precise, while this baseline obtains higher recall values (possibly related to oversegmentation). However, our method has the advantage of returning an output that can be easily converted into vectorial data.  As it can be observed in the visual comparisons in Section~\ref{ssec:remove_add_visual_comparison} (Figure~\ref{fig:visual_comparison_adding_buildings}), our method obtains building predictions with shapes that fit better to the ground truth, not oversegmenting. Also, in cases of objects with shared or very close boundaries, the buildings outlines are easily disentangled, while they cannot be recovered from the semantic segmentation results, since both objects are included in a single blob.

We also evaluate how accurate our method based on shape priors is in differentiating building shapes. To do so, we consider all the newly added buildings showing a considerable overlap (IoU $> 0.3$) with a building in the ground-truth map. Considering as classes the six basic primitive shapes, the predicted shapes obtains an accuracy of 90.0 \%. If we consider as classes the 18 shapes (therefore shape \emph{and} size of the object) an accuracy is 38.3 \% is reached. Most common errors are cases where the correct shape is predicted, but not the correct size.

For the evaluation of the geometrical accuracy of the new buildings, we use the average symmetric surface distance metric (ASSD). This metric computes the average distance between all the pixels in the boundary of the predicted object to the closest pixel in the boundary of the ground-truth object. A perfect building prediction will obtain an ASSD value of 0 (the lower the value the better it is). We have computed this metric for all the building predictions that have some overlap with the ground-truth. The average ASSD value for the predictions of the proposed method is 2.54 in the Tanzania dataset, while the method that add buildings based on semantic segmentation obtains an average ASSD value of 2.56.

\begin{figure}[!t]
\begin{center}
\begin{tabular}{cccc} 
\multicolumn{4}{c}{Example 1 - Correlation vs. MRF}\\
  \includegraphics[width=0.15\columnwidth]{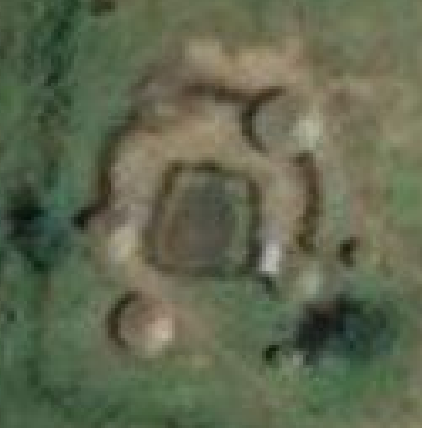} & 
  \includegraphics[width=0.15\columnwidth]{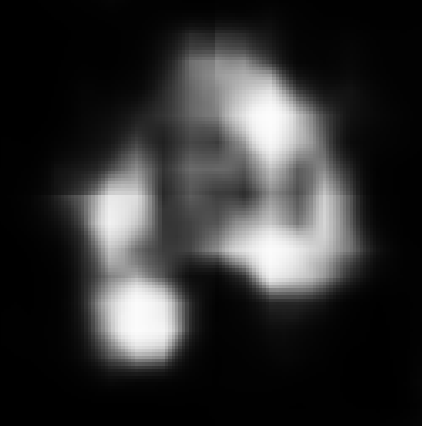} &  
  \includegraphics[width=0.15\columnwidth]{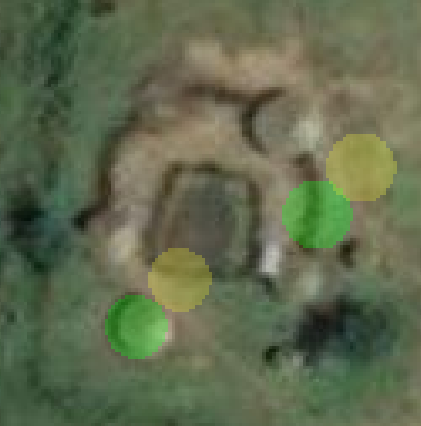} &
  \includegraphics[width=0.15\columnwidth]{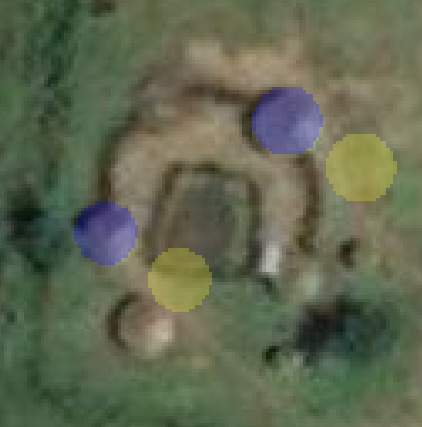} \\ 
  (a) Image & (b) CNN probability map & (c) \textcolor[rgb]{0,.8,0}{\texttt{CorrGroups}} & (d) \textcolor[rgb]{.25,.34,.55}{\texttt{MRFGroups}} \\
   \hline
  \multicolumn{4}{c}{Example 2 - Individual buildings vs. groups of buildings in the MRF}\\
  \includegraphics[width=0.15\columnwidth]{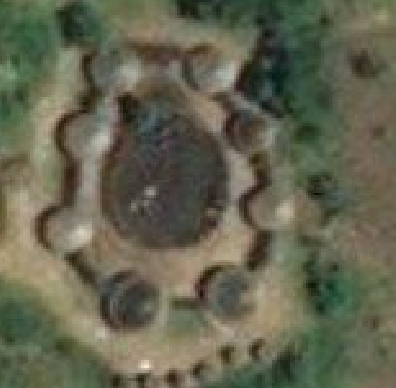} &
  \includegraphics[width=0.15\columnwidth]{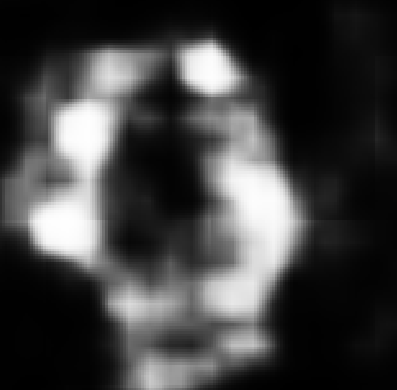} &
  \includegraphics[width=0.15\columnwidth]{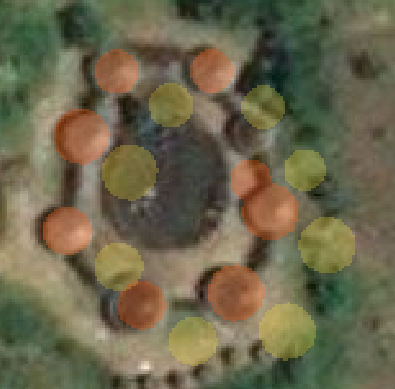} &
  \includegraphics[width=0.15\columnwidth]{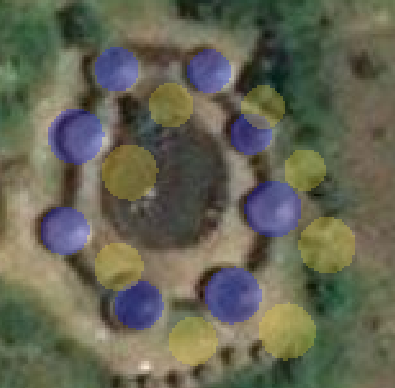} \\ 
  (e) Image & (f) CNN probability map & (g) \textcolor[rgb]{1,0.5,0}{\texttt{MRFBuildings}}  & (h) \textcolor[rgb]{.25,.34,.55}{\texttt{MRFGroups}} \\
      \hline
  \multicolumn{4}{c}{Example 3 - Correlation of individual buildings vs. MRF.}\\
  \includegraphics[width=0.15\columnwidth]{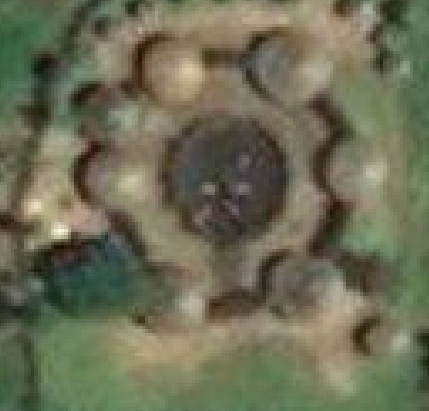} &
  \includegraphics[width=0.15\columnwidth]{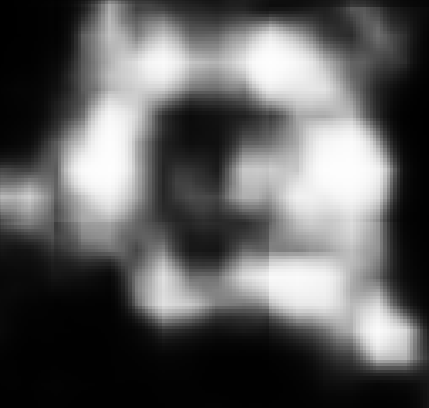} &
  \includegraphics[width=0.15\columnwidth]{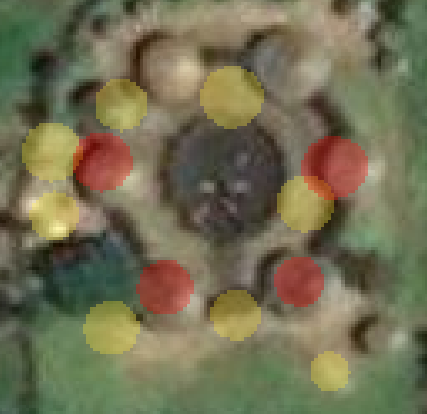} &
  \includegraphics[width=0.15\columnwidth]{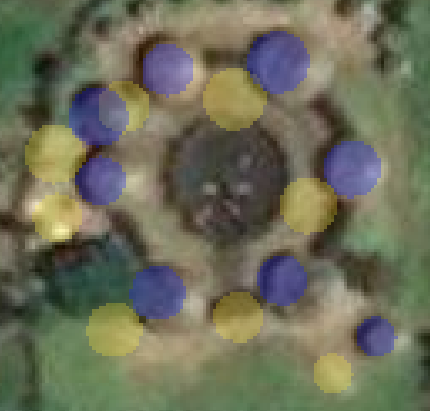} \\ 
  
  (i) Image  & (j) CNN probability map & (k) \textcolor[rgb]{1,0,0}{\texttt{CorrBuidings}}  & (l) \textcolor[rgb]{.25,.34,.55}{\texttt{MRFGroups}}\\

\hline
  \multicolumn{4}{c}{Example 4 - Absolute difference and  Mutual information}\\
  \includegraphics[width=0.15\columnwidth]{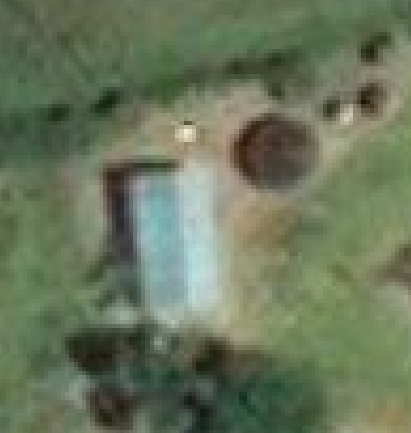} &
  \includegraphics[width=0.15\columnwidth]{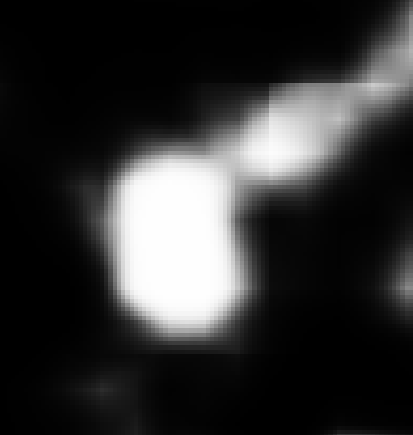} &
  \includegraphics[width=0.15\columnwidth]{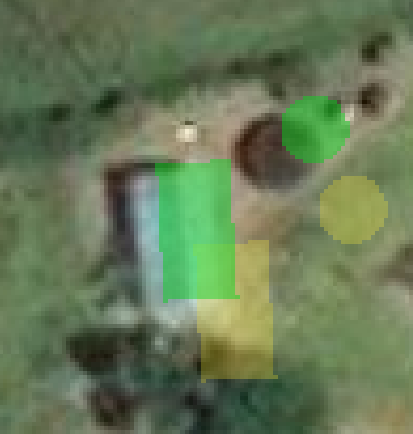} &
  \includegraphics[width=0.15\columnwidth]{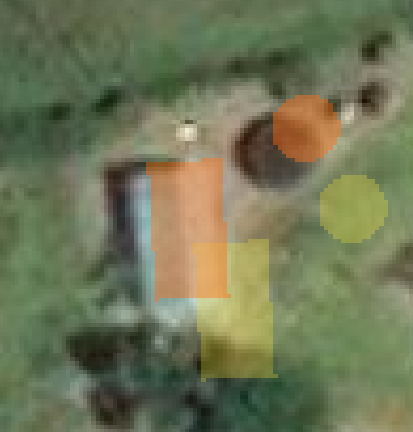} \\ 
  
  (m) Image  & (n) CNN probability map & (o) \textcolor[rgb]{0,.8,0}{\texttt{AbsDifference}}  & (p) \textcolor[rgb]{1,0.5,0}{\texttt{MutualInfo}}\\
 
\hline
  \multicolumn{4}{c}{Example 5 - Deformable registration}\\
  \includegraphics[width=0.15\columnwidth]{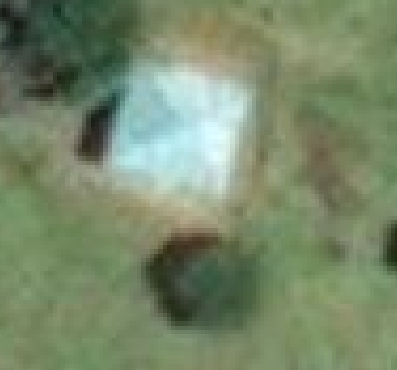} &
  \includegraphics[width=0.15\columnwidth]{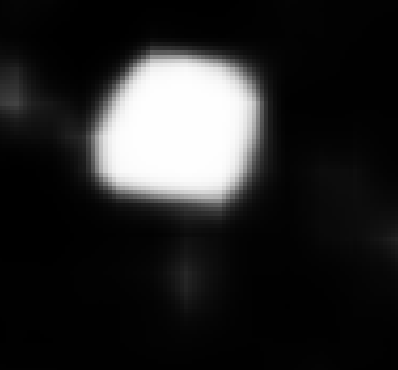} &
  \includegraphics[width=0.15\columnwidth]{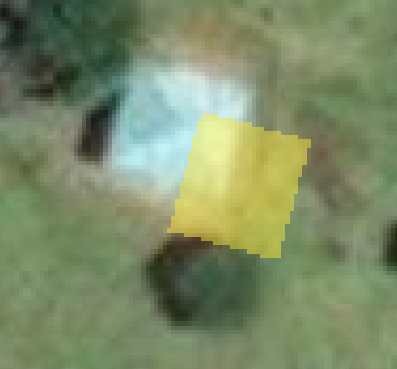} &
  \includegraphics[width=0.15\columnwidth]{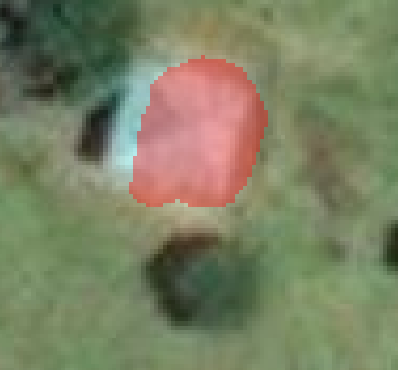} \\ 
  
  (q) Image  & (r) CNN probability map & (s) Original OSM  & (t) \textcolor[rgb]{1,0,0}{\texttt{DeformableReg}}\\
  
\end{tabular} 
\end{center}
\caption{Examples of alignment results (the original misaligned annotations are presented in yellow) from the Tanzania dataset.
\label{fig:alignment_examples3}}
\end{figure}

\subsection{Visual comparisons} 
\subsubsection{Alignment of footprints}
Figure~\ref{fig:alignment_examples3} presents five examples of groups of rural buildings from the Tanzania dataset. For each example, we show the image, the building probability maps obtained by the hypercolumn model, the original OSM annotations (in yellow) and the aligned annotations obtained by different methods (in other colors). For the proposed method, \texttt{MRFGroups}, only the alignment is performed and no removal / addition component is considered in the figure. 
\begin{itemize}
\item[-] Example 1 (first row). Figure~\ref{fig:alignment_examples3}c shows in green circles the aligned annotations obtained by \texttt{CorrGroups}. The alignment results obtained by our proposed method (\texttt{MRFGroups}), blue circles in Figure~\ref{fig:alignment_examples3}d, are more accurate, despite missing the bottom building, since the component that adds new building footprints was not used in this case. 
\item[-] Example 2 (second row). Figure~\ref{fig:alignment_examples3}g shows the alignment results obtained by the MRF applied on individual buildings (\texttt{MRFBuildings}, orange circles). One of the buildings was moved to an incorrect location because the values of the probability map are high in a location where there are no buildings. This does not happen in the case of the MRF applied over groups of buildings (\texttt{MRFGroups}, blue circles in Figure~\ref{fig:alignment_examples3}h) because we applied the prior knowledge that buildings that are spatially close should be registered with the same alignment vector. 
\item[-] Example 3 (third row). Figure~\ref{fig:alignment_examples3}k presents the results obtained by the alignment method that uses the correlation on individual buildings (\texttt{CorrBuildings}, red circles). We can observe that some building annotations are moved to the same location of high building probability values. The proposed \texttt{MRFGroups}, denoted by blue circles in Figure~\ref{fig:alignment_examples3}l, obtains a more desirable alignment, but still an inaccurate one (annotations are shifted to the left side of the buildings) because the building probability map itself is not accurate enough.
\item[-] Example 4 (forth row). 
Figures~\ref{fig:alignment_examples3}o and ~\ref{fig:alignment_examples3}p present the results obtained by \texttt{AbsDifference} and \texttt{MutualInfo}, respectively. We can observe that for both methods the two building annotations are not well aligned with the objects in the imagery. As for the previous example, this happens mainly because of inaccurate probability maps.
\item[-] Example 5 (fifth row). 
Figure~\ref{fig:alignment_examples3}t presents the result obtained by \texttt{DeformableReg} applied to correct the OSM annotation presented in Figure~\ref{fig:alignment_examples3}s. We can observe that the shape of the resulting annotation is inaccurate since it is registered to an object with an inaccurate shape in the building classification map. 

\end{itemize}

Although the proposed MRF based method is more robust to inaccurate building probability maps than the other alignment methods, the quality of the building probability map remains the main factor to compute accurate alignment vectors.

\begin{figure}[!t]
\begin{center}
\begin{tabular}{c|cc|cc} 
  \includegraphics[width=0.16\columnwidth,height=2.5cm]{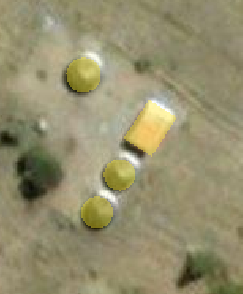} &
  \includegraphics[width=0.16\columnwidth,height=2.5cm]{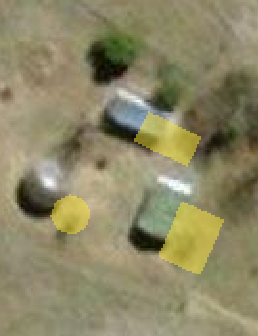} &
  \includegraphics[width=0.16\columnwidth,height=2.5cm]{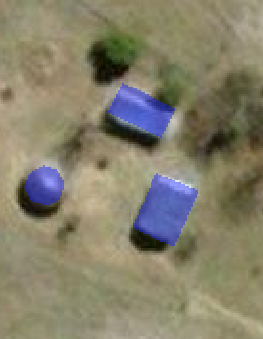} &
  \includegraphics[width=0.16\columnwidth,height=2.5cm]{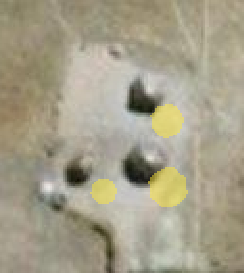} &
  \includegraphics[width=0.16\columnwidth,height=2.5cm]{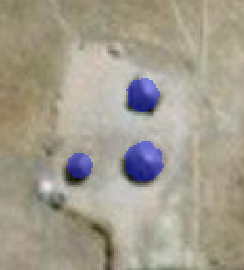} \\ 
  Original  & Original  & Aligned  & Original  & Aligned  \\
      annotations  &  annotations  &  annotations  &  annotations  & annotations\\
\end{tabular}
\caption{Examples of alignment results in the Zimbabwe dataset using \texttt{MRFGroups}.}\label{res:Zim}
\end{center}
\end{figure}

Figure~\ref{res:Zim} illustrates the alignment results for the proposed \texttt{MRFGroups} in three examples.  In the first case, no alignment is necessary, and \texttt{MRFGroups} result is equivalent to the original labels. In the two other cases, \texttt{MRFGroups} aligns the buildings correctly, and the removal and addition of footprints is not necessary. This is in line with expectations from this dataset, as we observe that the Zimbabwe dataset has better quality OSM annotations,  only requiring geometric alignment. Missing building annotations or incorrect annotations after alignment are rare. This is also reflected in Table~\ref{tab:results_zimbabwe}, in which the alignment of the original annotations considerably improved the performance, but the removal and addition of building annotations did not improve the final performance.

\subsubsection{Including footprint removals and additions}
\label{ssec:remove_add_visual_comparison}
\begin{figure}[!t]
\begin{center}
\begin{tabular}{ccccc} 
\hline
\multicolumn{5}{c}{Tanzania dataset}\\
  \includegraphics[width=0.16\columnwidth]{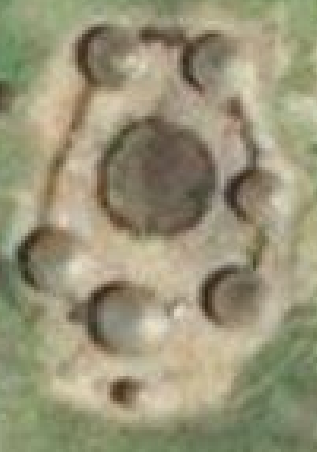} & 
  \includegraphics[width=0.16\columnwidth]{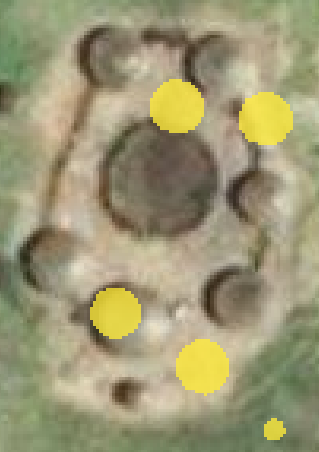} &  
  \includegraphics[width=0.16\columnwidth]{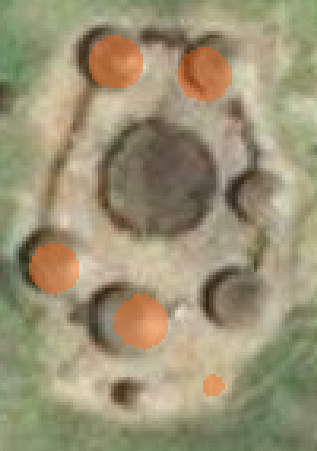} &
  \includegraphics[width=0.16\columnwidth]{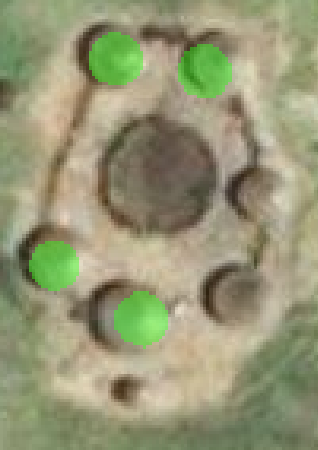} &
  \includegraphics[width=0.16\columnwidth]{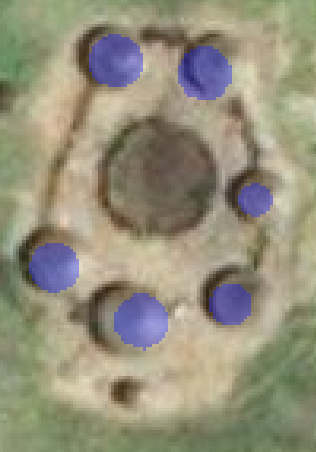} \\ 
  \multirow{2}{*}{(a) Image} & (b) Original  & (c) \texttt{MRFGroups}  & (d) \texttt{MRFGroups}+ & (e) \texttt{MRFGroups}+ \\
                            &  annotations &   &  removal & removal+addition\\
   \hline
   \multicolumn{5}{c}{Zimbabwe dataset}\\
   \includegraphics[width=0.16\columnwidth]{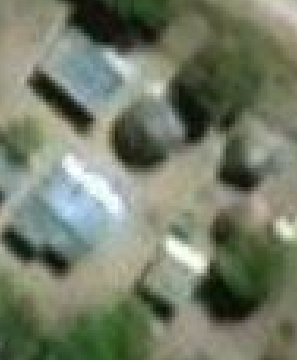} &
   \includegraphics[width=0.16\columnwidth]{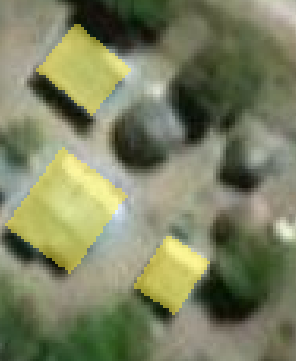} &
   \includegraphics[width=0.16\columnwidth]{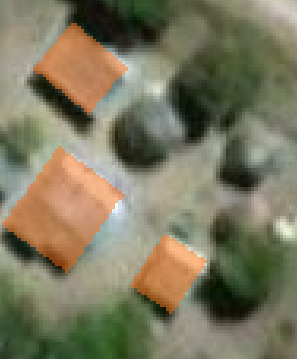} &
   \includegraphics[width=0.16\columnwidth]{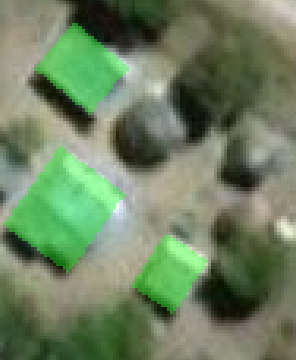} &
   \includegraphics[width=0.16\columnwidth]{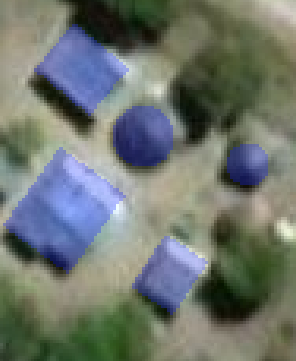} \\ 
  \multirow{2}{*}{(f) Image} & (g) Original  & (h) \texttt{MRFGroups}  & (i) \texttt{MRFGroups}+ & (j) \texttt{MRFGroups}+ \\
                            &  annotations &   & removal & removal+addition
  
\end{tabular} 
\end{center}
\caption{Results of our method (the original misaligned annotations are presented in yellow) for the Tanzania and Zimbabwe dataset. 
  \label{fig:tanzania_zimbabwe_examples}}
\end{figure}

Figure~\ref{fig:tanzania_zimbabwe_examples} presents results of the methods for alignment (orange), removal of incorrect annotations (green) and addition of new annotations (blue) in the Tanzania and Zimbabwe datasets. For Tanzania dataset example, on the top row, an incomplete set of annotations (Figure~\ref{fig:tanzania_zimbabwe_examples}b) is first geometrically aligned so that the large buildings correspond to structures in the image (Figure~\ref{fig:tanzania_zimbabwe_examples}c); then, the small structure at the bottom is removed, since there is no evidence that a small building would be located there (Figure~\ref{fig:tanzania_zimbabwe_examples}d). One could argue that the removed building corresponds to a small structure at the bottom, but given the relative configuration of the annotations, this is against the image evidence learned by the CNN model. Finally, the second CNN adding new footprints succeeds in adding the two missing large buildings in the right side (Figure~\ref{fig:tanzania_zimbabwe_examples}e). For the example from the Zimbabwe dataset (Figure~\ref{fig:tanzania_zimbabwe_examples}f), the original OSM annotations  (Figure~\ref{fig:tanzania_zimbabwe_examples}g) are already well aligned. As a consequence, the alignment correction (Figure~\ref{fig:tanzania_zimbabwe_examples}h) and  the removal of incorrect annotations (Figure~\ref{fig:tanzania_zimbabwe_examples}i) do not change the location of the original annotations. However, two new footprints of missing buildings are correctly added using the second CNN (Figure~\ref{fig:tanzania_zimbabwe_examples}j).

\begin{figure}[!t]
\begin{center}
\begin{tabular}{cccc} 
  \includegraphics[width=0.2\columnwidth]{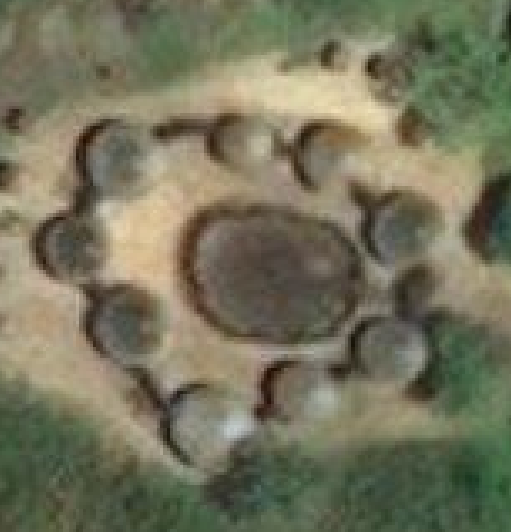} & 
  \includegraphics[width=0.2\columnwidth]{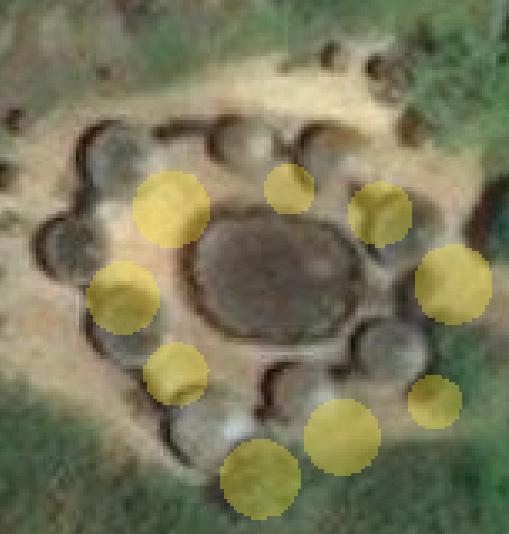} &  
  \includegraphics[width=0.2\columnwidth]{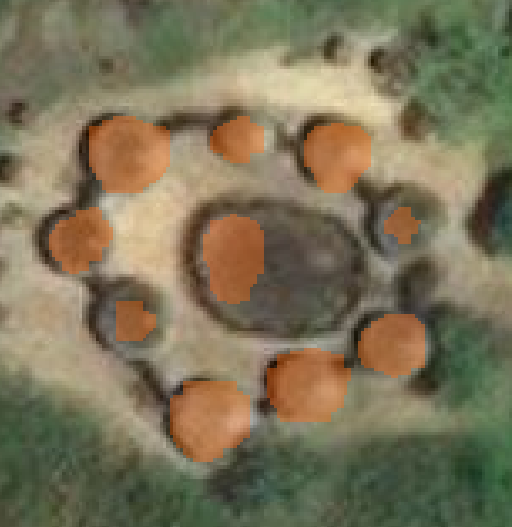} &
  \includegraphics[width=0.2\columnwidth]{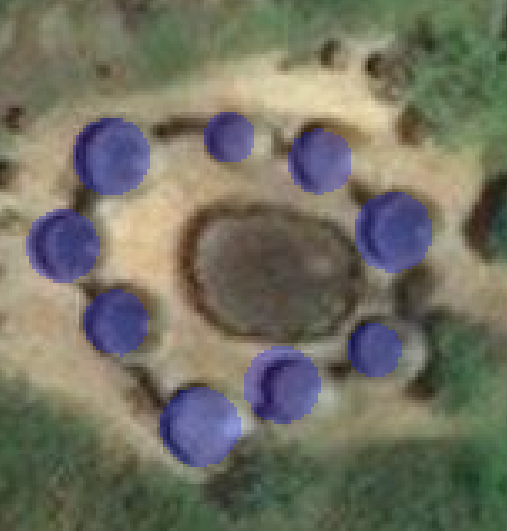} \\ 
  \multirow{2}{*}{(a) Image} & \multirow{2}{*}{(b) Original OSM}  & (c) \texttt{Semantic}  & (d) \texttt{MRFGroups}+ \\
                            &   &  \texttt{segmentation~\cite{Maggiori_2017cnn}} & removal+addition  
  
\end{tabular} 
\end{center}
\caption{Results of our method compared with semantic segmentation~\cite{Maggiori_2017cnn}: a) Imagery of groups of buildings b) Original OSM annotations (yellow circles) c) Results obtained by using a CNN model trained for building segmentation (orange circles) and d) Annotations, in blue circles,  obtained using the propose method (MRF alignment followed by removal and addition of annotations) 
	\label{fig:example_automatic_classification}}
\end{figure}

Figure~\ref{fig:example_automatic_classification} compares the results obtained by our proposed method (\texttt{MRFGroups} followed by the removal and addition of building annotations) with the result of a CNN-based method trained for building segmentation~\cite{Maggiori_2017cnn}. We can observe that, despite detecting most buildings, the prediction of the CNN segmentation model is not precise, containing several false positive pixels, while our proposed method obtains a better result, more coherent with the shapes of the buildings to be detected.

\begin{figure}[!t]
\begin{center}
\begin{tabular}{ccc} 

\multicolumn{3}{c}{Example 1}\\
  \includegraphics[width=0.25\columnwidth]{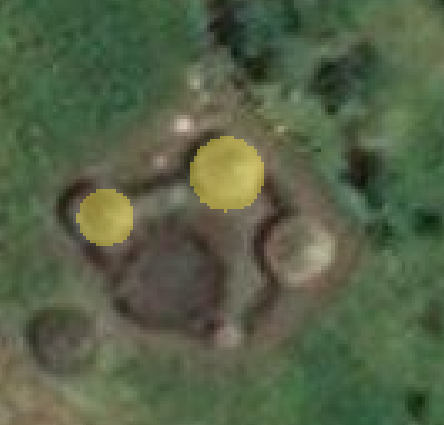} & 
  \includegraphics[width=0.25\columnwidth]{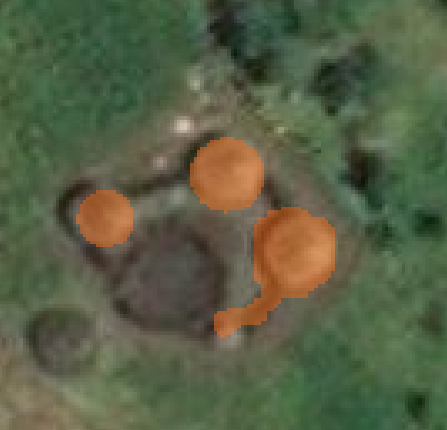} &
  \includegraphics[width=0.25\columnwidth]{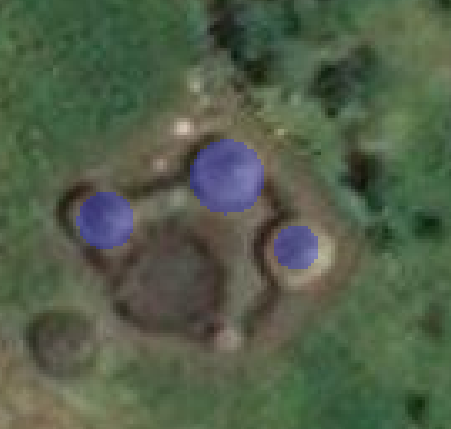} \\
  (a) \texttt{MRFGroups}+  & (b) \texttt{Semantic}  & (c) \texttt{MRFGroups}+ \\
                             removal &  \texttt{segmentation} & removal+addition\\
  \hline
  \multicolumn{3}{c}{Example 2}\\
  \includegraphics[width=0.25\columnwidth]{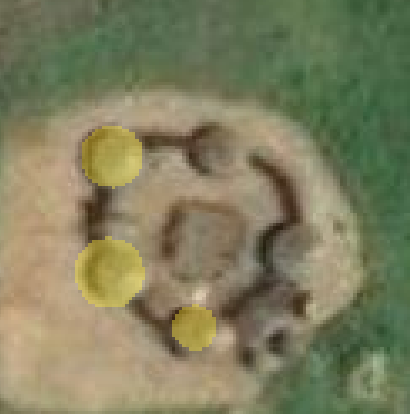} & 
  \includegraphics[width=0.25\columnwidth]{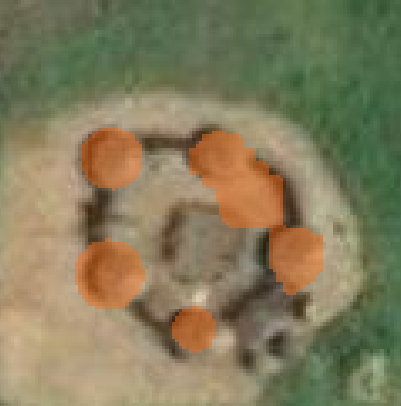} &
  \includegraphics[width=0.25\columnwidth]{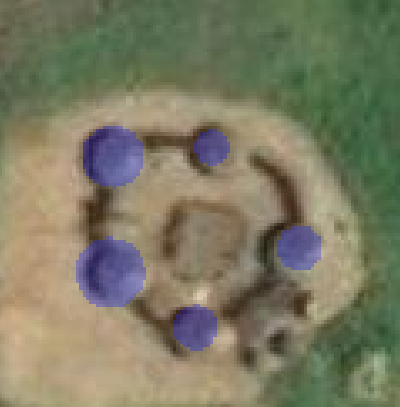} \\
  (d) \texttt{MRFGroups}+  & (e) \texttt{Semantic}  & (f) \texttt{MRFGroups}+ \\
                             removal &  \texttt{segmentation} & removal+addition\\
  \hline
  \multicolumn{3}{c}{Example 3}\\
  \includegraphics[width=0.25\columnwidth]{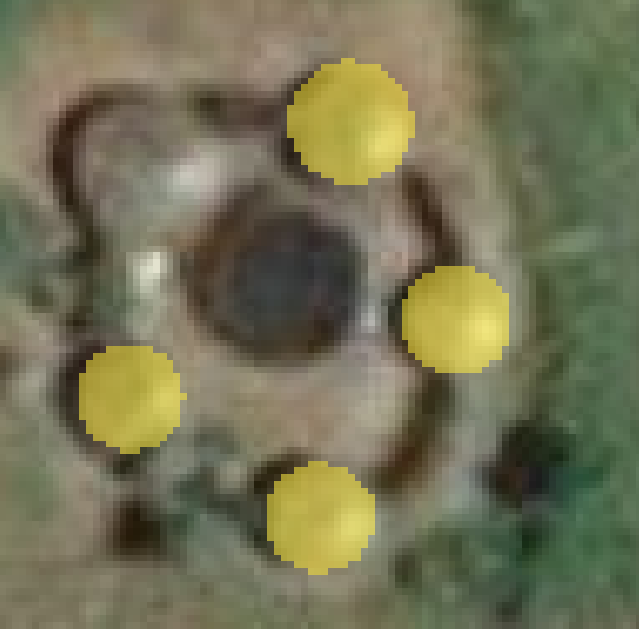} & 
  \includegraphics[width=0.25\columnwidth]{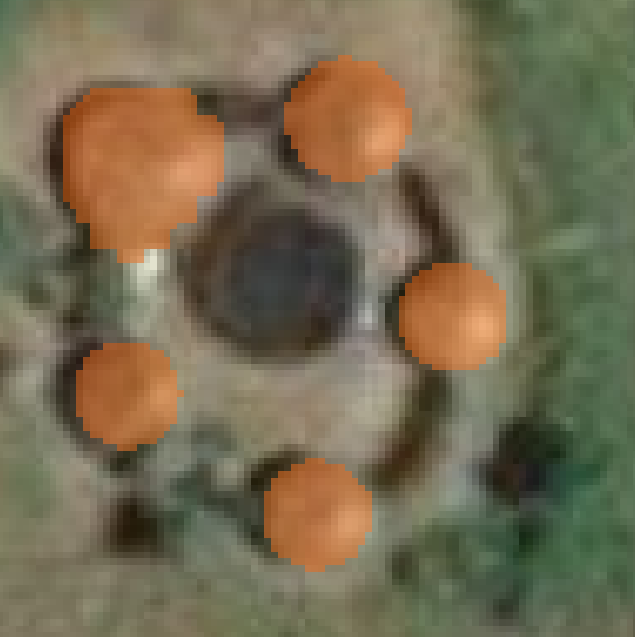} &
  \includegraphics[width=0.25\columnwidth]{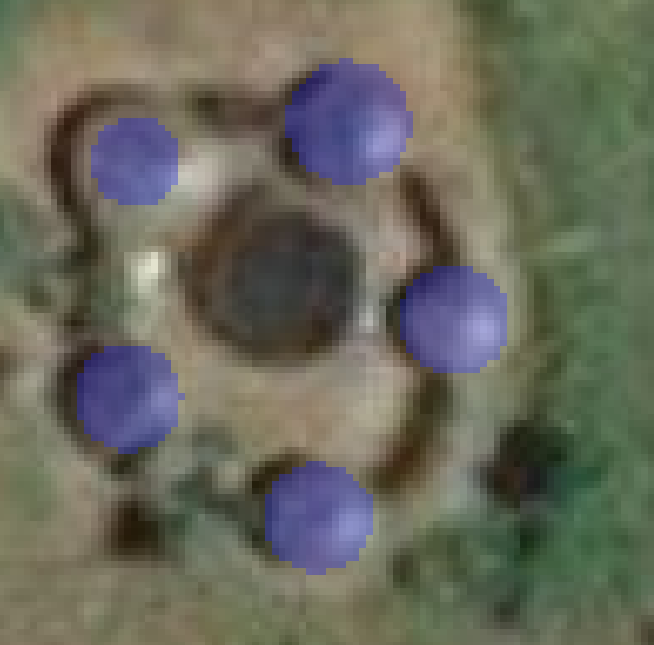} \\
  (g) \texttt{MRFGroups}+  & (h) \texttt{Semantic}  & (i) \texttt{MRFGroups}+ \\
                             removal &  \texttt{segmentation} & removal+addition\\
\end{tabular} 
\end{center}
\caption{Visual comparison of two methods for adding new building annotations, after the alignment and removal of annotations. 1) Add new buildings using the semantic segmentation method proposed in~\cite{Maggiori_2017cnn} and 2) the proposed method based on shape priors.
\label{fig:visual_comparison_adding_buildings}}
\end{figure}

Figure~\ref{fig:visual_comparison_adding_buildings} shows three examples of comparisons of the results of adding buildings using a semantic segmentation method~\cite{Maggiori_2017cnn} and our proposed method for adding building annotations, based on shape priors. The shape of the output of the semantic segmentation method can be very irregular, while our proposed methods obtain predictions that fits better to the ground truth (see examples 1 and 2). In some cases, the prediction of the semantic segmentation method can obtain higher values of IoU with the ground truth than our proposed method since it tends to predict more pixels as buildings (oversegmentation). However, it can also obtain some undesirable results like in Figure~\ref{fig:visual_comparison_adding_buildings}e. Overall, the proposed method leads to a more precise outlining of buildings, easily exportable to vector footprints, and also can disambiguate effectively with polygons with very close boundaries.

\section{Conclusion}
\label{sec:conclusions}

We presented a methodology for correcting rural building annotations in OpenStreetMap. Our methodology consists of three steps: alignment of the original annotations, removal of incorrect annotations, and addition of new annotations of buildings that appear for the first time in the updated imagery. In order to solve the problem of misaligned OSM annotations, we proposed an MRF-based method that encodes the dependency of the alignment vectors of neighboring buildings and maximizes the correlation of aligned annotations and a building probability map learned by a fully convolutional neural network. We used the evidence provided by a building probability map to remove annotations of buildings that no longer exist in the updated imagery. In order to add new building annotations, we learn a second CNN model that predicts building annotations with predefined shapes candidates. We evaluated our methodology in a region of Tanzania that contains misaligned and incomplete/inaccurate annotations and in a region in Zimbabwe that contains mostly misaligned annotations. We observed that the alignment process drastically improves the accuracy of the annotations in the two evaluated datasets. We observed, specially in the Tanzania dataset, that the proposed method for the removal of annotations improves the precision of the annotations and the proposed method for the addition of new annotations considerably improves the recall of the annotations. The proposed methodology will be helpful to reduce the large human effort required to correct existing rural building OSM annotations. 
As future work, we plan to improve the building delineation results by combining building probability maps learned by CNNs, graph-based segmentation methods, and shape priors.

\section*{Acknowledgment}
This research was funded by FAPESP (grant 2016/14760-5, 2017/10086-0 and 2014/12236-1), the CNPq (grant 302970/2014-2) and by the Swiss National Science Foundation (grant PP00P2-150593).

\section*{References}
\bibliography{refs}

\end{document}